\documentclass[conference]{IEEEtran}
\IEEEoverridecommandlockouts
\usepackage[english]{babel}
\usepackage[utf8]{inputenc}
\usepackage{cite}
\usepackage{amsmath,amssymb,amsfonts}
\usepackage{algorithmic}
\usepackage{graphicx}
\usepackage{textcomp}
\usepackage{xcolor}
\usepackage[utf8]{inputenc}

\def\BibTeX{{\rm B\kern-.05em{\sc i\kern-.025em b}\kern-.08emT\kern-.1667em\lower.7ex\hbox{E}\kern-.125emX}}

\begin{document}
	
	\title{Evaluation Metrics for Unsupervised Learning Algorithms}
	\author{
		\IEEEauthorblockN{Julio-Omar Palacio-Niño}
		\IEEEauthorblockA{
			\textit{Dept. Computer Science and Artificial Intelligence} \\
			\textit{Universidad de Granada}\\
			Granada, Spain \\
			jopalacion@correo.ugr.es
		}
		\and
		\IEEEauthorblockN{Fernando Berzal}
		\IEEEauthorblockA{
			\textit{Dept. Computer Science and Artificial Intelligence} \\
			\textit{Universidad de Granada}\\
			Granada, Spain \\
			berzal@acm.org
		}
	}

	\maketitle
	
\begin{abstract}
Determining the quality of the results obtained by clustering techniques is a key issue in unsupervised machine learning. Many authors have discussed the desirable features of good clustering algorithms. However, Jon Kleinberg established an impossibility theorem for clustering. As a consequence, a wealth of studies have proposed techniques to evaluate the quality of clustering results depending on the characteristics of the clustering problem and the algorithmic technique employed to cluster data.\\
\end{abstract}
	

\section{Introduction}

Machine learning techniques are usually classified into supervised and unsupervised techniques. Supervised machine learning starts from prior knowledge of the desired result in the form of labeled data sets, which allows to guide the training process, whereas unsupervised machine learning works directly on unlabeled data. In the absence of labels to orient the learning process, these labels must be ``discovered'' by the learning algorithm.\cite{cord_unsupervised_2008}\\
	
In this technical report, we discuss the desirable features of good clustering results, recall Kleinberg's impossibility theorem for clustering, and describe a taxonomy of evaluation criteria for unsupervised machine learning. We also survey many of the evaluation metrics that have been proposed in the literature. We end our report by describing the techniques that can be used to adjust the parameters of clustering algorithms, i.e. their hyperparameters.\\

\section{Formal Limitations of Clustering}
	
From an intuitive point of view, the clustering problem has a very clear goal; namely, properly clustering a set of unlabeled data. Despite its intuitive appeal, the notion of ``cluster'' cannot be precisely defined, hence the wide range of clustering algorithms that have been proposed.\cite{Estivill-Castro2002}\\
	
\subsection{Desirable Features of Clustering}
	
Jon	Kleinberg proposes three axioms that highlight the characteristics that a grouping problem should exhibit and can be considered ``good'', independently of the algorithm used to find the solution. These axioms are scale invariance, consistency, and wealth \cite{kleinberg_impossibility_2002}, which are explained in more detail below.\\
	
A grouping function is defined as a set of $S$ of $x \geq 2$ points and the distances between pairs of points. The set of points is $S=\{1,2,...,n\}$ and the distance between points is given by the distance function $d(i,j)$, where $i,j \in S$. The distance function $d$ measures the dissimilarity between pairs of points. For instance, the Euclidean, Manhattan, Chebyshev, and Mahalanobis distances can be used, among many others. Alternatively, a similarity function might also be used.\\		
	
\subsubsection{Scale Invariance}
	
The first of Kleinberg's axioms states that $f(d)=f(\alpha \cdot d)$ for any distance function $d$ and any scaling factor $\alpha>0$.\cite{kleinberg_impossibility_2002}\\
	
This simple axiom indicates that a clustering algorithm should not modify its results when all distances between points are scaled by the factor determined by a constant $\alpha$.\\

\subsubsection{Richness}

A clustering process is considered to be rich when every partition of $S$ is a possible result of the clustering process. If the use $Range(f)$ to denote the set of all $\Gamma$ partitions so that $f(d)=\Gamma$ for some distance function $d$, then $Range(f)$ is equal to the set of all $S$ partitions.\cite{kleinberg_impossibility_2002}\\
	
This means that the the clustering function must be flexible enough to produce any arbitrary partition/clustering of the input data set.\\
	
\subsubsection{Consistency}
	
Let $d$ and $d'$ be two distance functions. If, for every pair $(i,j)$ belonging to the same cluster, $d(i,j) \geq d'(i,j)$, and for every pair $(i,j)$ belonging to different clusters, $d(i,j) \leq d'(i,j)$, then $f(d)=f(d')$.\cite{kleinberg_impossibility_2002}\\
	
A clustering process is ``consistent'' when the clustering results do not change if the distances within clusters decrease and/or the distances between clusters increase.\\

\subsection{An Impossibility Theorem for Clustering}\
	
Given the above three axioms, Kleinberg proves the following theorem: For every $n>=2$, there is no clustering function $f$ that satisfies scale invariance, richness, and consistency.\cite{kleinberg_impossibility_2002}\\
	
Determining a ``good'' clustering is not a trivial problem. It is impossible for any clustering procedure to be able to satisfy all three axioms. Practical clustering algorithms must trade-off the desirable features of clustering results.\\
	
Since the three axioms cannot hold simultaneously, clustering algorithms can be designed to violate one of the axioms while sarisfying the other two. Kleinberg illustrates this point by describing three variants of single-link clustering (an agglomerative hierarchical clustering algorithm): \cite{kleinberg_impossibility_2002}\\
	
\begin{itemize}

\item
$k$-cluster stopping condition: Stop merging clusters when we have $k$ clusters (violates the richness axiom, since the algorithm would never return a number of clusters different to $k$).\\

\item
Distance-$r$ stopping condition: Stop merging clusters when the nearest pair of clusters are farther than $r$ (violates scale invariance given that every cluster will contain a single instance when $\alpha$ is large, whereas a single cluster will contain all data when $\alpha \rightarrow 0$).\\

\item
Scale-$\epsilon$ stopping condition: Stop merging clusters when the nearest pair of clusters are farther than a fraction $\epsilon$ of the maximum pairwise distance $\Delta$ (scale invariance is satisfied, yet consistency is violated).\\

\end{itemize}
	
Clustering algorithms can often satisfy the properties of scale invariance and consistency by relaxing their richness (e.g. whenever the number of clusters is established beforehand). As we have seen, some algorithms can even be customized to satisfy two out of three axioms by relaxing the third one (e.g. simple linkage with different stopping criteria).\\

\section{Methods for Cluster Evaluation}
	
Evaluating the results of a clustering algorithm is a very important part of the process of clustering data. In supervised learning,``the evaluation of the resulting classification model is an integral part of the process of developing a classification model and there are well-accepted evaluation measures and procedures'' \cite{tan_introduction_2005}. In unsupervised learning, because of its very nature, cluster evaluation, also known as cluster validation, is not as well-developed.\cite{tan_introduction_2005} \\
	
In clustering problems, it is not easy to determine the quality of a clustering algorithm. This gives rise to multiple evaluation techniques. Quite often, the evaluation process includes a notorious particularity: the way the measurement is performed depends on the algorithm used to obtain the clustering results.\\
	
When analyzing clustering results, several aspects must be taken into account for the validation of the algorithm results\cite{tan_introduction_2005}:\\
	
\begin{itemize}
		\item Determining the clustering tendency in the data (i.e. whether non-random structure really exists).\\
		\item Determining the correct number of clusters.\\
		\item Assessing the quality of the clustering results without external information.\\
		\item Comparing the results obtained with external information.\\
		\item Comparing two sets of clusters to determine which one is better.\\
\end{itemize}

The first three issues are addressed by \textbf{internal or unsupervised validation}, because there is no use of external information. The fourth issue is resolved by \textbf{external or supervised validation}. Finally, the last issue can be addressed by both supervised and unsupervised validation techniques.\cite{tan_introduction_2005}.\\
	
\begin{figure}[htbp]
		\centerline
		{
			\includegraphics[scale=0.3]{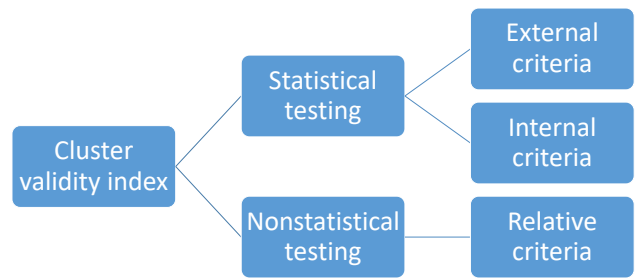}
		}
		\caption{Taxonomy of clustering evaluation methods (adaptation)\cite{gan_data_2007}.}
		\label{figura1}
	\end{figure}

Gan et al. \cite{gan_data_2007} propose a taxonomy of evaluation techniques that comprises both internal and external validation approaches (see Figure \ref{figura1}).\\
	
\subsection{Null Hypothesis Testing}
	
One of the desirable characteristics of a clustering process is to show whether data exhibits some tendency to form actual clusters. From a statistical point of view, a feasible approach consists of testing whether data exhibits random behavior or not \cite{halkidi_clustering_2001}. In this context, the null hypothesis testing can be used: A null hypothesis $H_0$ is assumed to be true until evidence suggests otherwise. In this case, the null hypothesis is the randomness of data and, when the null hypothesis is rejected, we assume that the data is significantly unlikely to be random. \cite{gan_data_2007}.\\

One of the difficulties of null hypothesis testing in this context is determining the statistical distribution under which the randomness hypothesis can be rejected. Jain and Dubes propose three alternatives \cite{jain_algorithms_1988}:\\
	
\begin{itemize}
  \item Random plot hypothesis $H_0$: All proximity matrices of order $n \times n$ are equally likely.\\
  \item Random label hypothesis $H_0$: All permutations of labels on $n$ objects are equally likely.\\
  \item Random position hypothesis $H_0$: All sets of $n$ locations in some region of a $d$-dimensional space are equally likely.	\\
\end{itemize}

Statistical techniques such as Monte Carlo analysis and bootstrapping can be used to determine the clustering tendency in data \cite{jain_algorithms_1988}.\\

\section{Internal validation}

Internal validation methods make it possible to establish the quality of the clustering structure without having access to external information (i.e. they are based on the information provided by data used as input to the clustering algorithm).\\

\begin{figure}[htbp]
		\centerline
		{
			\includegraphics[scale=0.8]{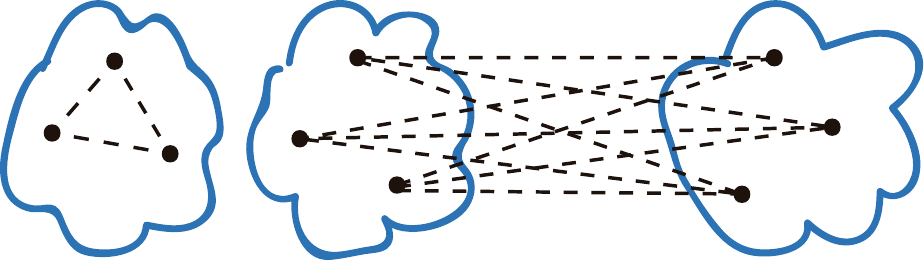}
		}
		\caption{Illustration of cohesion and separation \cite{tan_introduction_2005}.}
		\label{figura2}
	\end{figure}

In general, two types of internal validation metrics can be combined: cohesion and separation measures. Cohesion evaluates how closely the elements of the same cluster are to each other, while separation measures quantify the level of separation between clusters (see Figure \ref{figura2}). These measures are also known as internal indices because they are computed from the input data without any external information \cite{tan_introduction_2005}. Internal indices are usually employed in conjunction with two clustering algorithm families: hierarchical clustering algorithms and partitional algorithms.\cite{gan_data_2007}.\\
	
Internal validation is used when there is no additional information available. In most cases, the particular metrics used by the evaluation methods are the same metrics that the clustering algorithm tries to optimize, which can be counterproductive in determining the quality of a clustering algorithm and deliver unfair validation results. On the other hand, in the absence of other sources of information, these metrics allow different algorithms to be compared under the same evaluation criterion \cite{aggarwal_data_2015}, yet care must be taken not to report biased results.\\
	
\begin{figure}[htbp]
			\centerline
			{
				\includegraphics[scale=0.3]{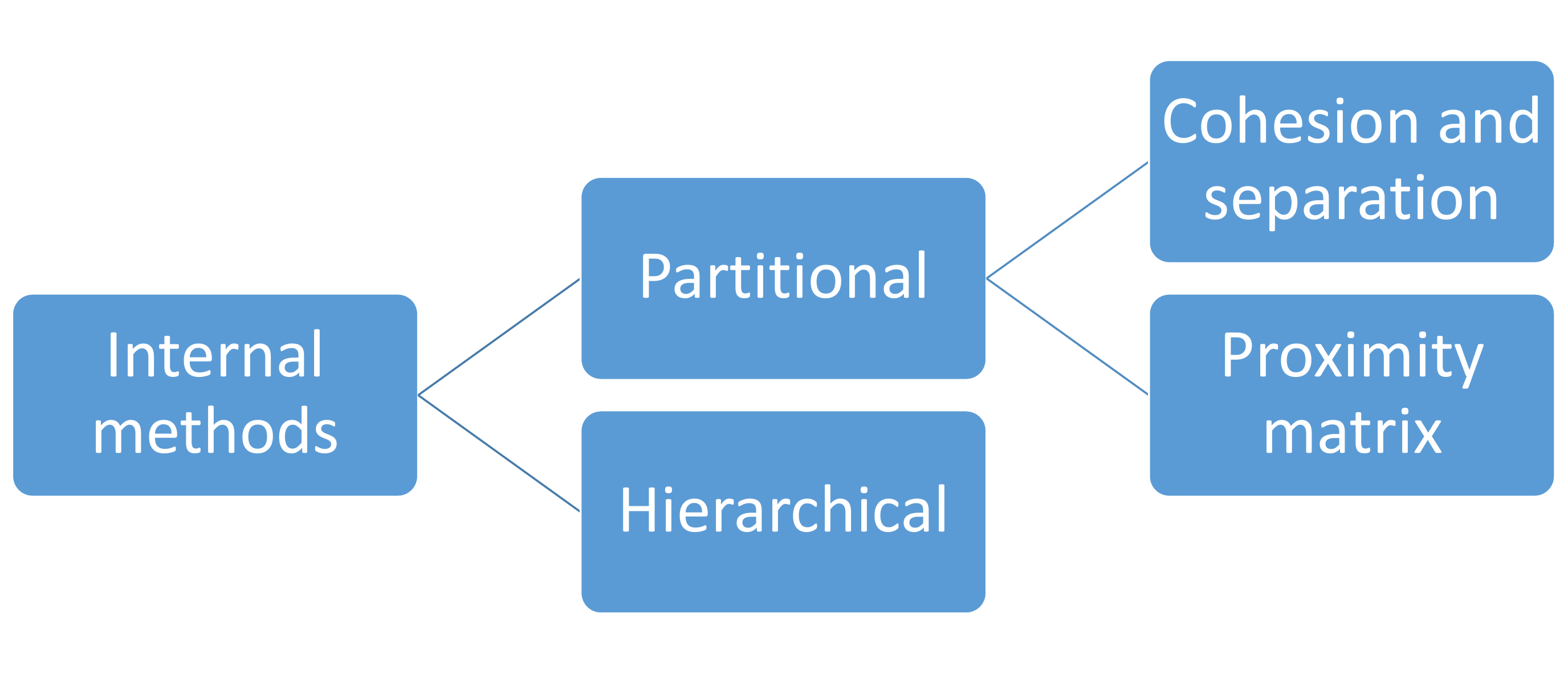}
			}
			\caption{Internal validation methods \cite{tan_introduction_2005}.}
			\label{figura3}
		\end{figure}

Internal evaluation methods are commonly classified according to the type of clustering algorithm they are used with. For partitional algorithms, metrics based on the proximity matrix, as well as metrics of cohesion and separation, such as the silhouette coefficient, are often used. For hierarchical algorithms, the cophenetic coefficient is the most common (see Figure \ref{figura3}).\\
	
\subsection{Partitional Methods}

	
Several of the measures employed by internal cluster validations methods are based on the concepts of cohesion and separation (see Figure \ref{figura2}). In general, the internal validation value of a set of $K$ clusters can be decomposed as the sum of the validation values for each cluster \cite{tan_introduction_2005}:\\
	
		\begin{equation*}
			general~validity=\sum_{i=1}^K w_i~validity(C_i)\label{eq1}
		\end{equation*}
	
This measure of validity can be cohesion, separation, or some combination of both. Quite often, the weights that appear in the previous expression correspond to cluster size.\\

The individual measures of cohesion and separation are defined as follows\cite{tan_introduction_2005}:\\
	
		\begin{equation*}
			cohesion(C_i) =\sum_{x \in C_i, y \in C_i} proximity(x,y)\label{eq2}
		\end{equation*}	

		\begin{equation*}
			separation(C_i, C_j) =\sum_{x \in C_i, y \in C_j} proximity(x,y)\label{eq3}
		\end{equation*}	

Cohesion is measured within a cluster (an intra-cluster metric), whereas separation is measured between clusters (an inter-cluster measure). Both are based on a proximity function that determines how similar a pair of examples are (similarity, dissimilarity and distance functions can be used). These metrics can also be defined for prototype-based clustering techniques, where proximity is measured from data examples to cluster centroids or medoids.\\

It should be noted that the cohesion metric defined above is equivalent to the cluster SSE [Sum of Squared Errors], also known as SSW [Sum of Squared Errors Within Cluster], when the proximity function is the squared Euclidean distance \cite{tan_introduction_2005}:\\
	
		\begin{equation*}
			SSE(C_i) = \sum_{x \in C_i} d(c_i,x)^2 = \frac{1}{2m_i}\sum_{x \in C_i}\sum_{y \in C_i} d(x,y)^2\label{eq4}
		\end{equation*}
		
where $x$ is an example in the cluster, $c_i$ is a cluster representative (e.g. its centroid) and $m_i$ is the number of examples in the cluster $C_i$.\\
	
When using the SSE metric, small values indicate a good cluster quality. Obviously, this metric is minimized in those clusters that were built from SSE-optimization-based algorithms such as k-means, but is clearly suboptimal for clusters detected using other techniques, such as density-based algorithms (e.g. DBSCAN) \cite{aggarwal_data_2015}.\\
	
Likewise, we can maximize the distance between clusters using a separation metric. This approach leads to the between group sum of squares, or SSB \cite{tan_introduction_2005}:\\
	
		\begin{equation*}
			SSB=\sum_{i=1}^K~m_i~ d(c_i,c)^2 = \frac{1}{2K} \sum_{i=1}^K \sum_{j=1}^K \frac{m}{K} d(c_i,c_j)^2\label{eq5}
		\end{equation*}
		
where $c_i$ is the mean of the $i^{th}$ cluster and $c$ is the overall mean  \cite{tan_introduction_2005}. Unlike the SSE metric, a good cluster quality is given by the high SSB values. As before, SSB is biased in favor of algorithms based on maximizing the separation distances among cluster centroids.\cite{aggarwal_data_2015}.\\
	
As mentioned above, a clustering is considered to be good when it has a high separation between clusters and a high cohesion within clusters \cite{handl_computational_2005}. Instead of dealing with separate metrics for cohesion and separation, there are several metrics that try to quantify the level of separation and cohesion in a single measure \cite{zhao_cluster_2012}:\\
	
\begin{itemize}

\item
The Calisnki-Harabasz coefficient, CH, also known as the variance ratio criterion, is a measure based on the internal dispersion of clusters and the dispersion between clusters. We would choose the number of clusters that maximizes the CH value for $M$ clusters \cite{calinski_dendrite_1974}:

			\begin{equation*}
				CH=\frac{\frac{SSB_M}{(M-1)}}{\frac{SSE_M}{(M)}}\label{eq6}
			\end{equation*}\\
			
\item
The Dunn index is the ratio of the smallest distance between data from different clusters and the largest distance between clusters. Again, this ratio should be maximized \cite{dunn_well-separated_1974}:
			\begin{equation*}
				D=\min_{1<i<k}\left\{\min_{1<j<k,i\neq j}\left\{\frac{\delta\left(C_i,C_j \right)}{\max_{1<l<k}\left \{ \Delta \left ( C_l \right ) \right \}}\right\}\right\} \label{eq7}
			\end{equation*}
			
			\begin{equation*}
				\Delta(C_i)=\max_{x,y\in c_i}\{ d ( x , y)\} \label{eq8}
			\end{equation*}	
			
			\begin{equation*}
				\delta(C_i,C_j)=\min_{x\in C_i,y \in C_j}\{ d(x,y) \} \label{eq9}
			\end{equation*}\\
			
\item
The Xie-Beni score was designed for fuzzy clustering, but it can applied to hard clustering. As the previous coefficients, it is a ratio whose numerator estimates the level of compaction of the data within the same cluster and whose denominator estimates the level of separation of the data from different clusters \cite{xie_validity_1991}:
			
			\begin{equation*}
				XB=\frac{\sum_{i=1}^{N}\sum_{k=1}^{M}u_{ik}^{2}\||x_i - C_k \||^2}{N_{t\neq s}\min\{\||C_t - C_s\||^2\}} \label{eq10}
			\end{equation*}\\	
			
\item
The Ball-Hall index is a dispersion measure based on the quadratic distances of the cluster points with respect to their centroid \cite{ball_isodata_1965}:
			
			\begin{equation*}
				BH=\frac{SSE_M}{M} \label{eq11}
			\end{equation*}\\
			
\item
The Hartigan index is based on the logarithmic relationship between the sum of squares within the cluster and the sum of squares between clusters \cite{hartigan_clustering_1975}.
			
			\begin{equation*}
				H=\log \left ( \frac{SSB_M}{SSE_M} \right ) \label{eq12}
			\end{equation*}\\
			
\item
The Xu coefficient takes into account the dimensionality $D$ of the data, the number $N$ of data examples, and the sum of squared errors $SSE_M$ form $M$ clusters \cite{xu_bayesian_1997}:
			
			\begin{equation*}
				Xu=D\log_2\left ( \sqrt{\frac{SSE_M}{DN^2}} \right )+\log M \label{eq13}
			\end{equation*}	\\	
\item
The silhouette coefficient is the most common way to combine the metrics of cohesion and separation in a single measure. Computing the silhouette coefficient at a particular point consists of the following three steps \cite{tan_introduction_2005}:
	
	\begin{enumerate}
		\item 	
        For each example, the average distance $a(i)$ to all the examples in the same cluster is computed:
		
		\begin{equation*}
			a(i)=\frac{1}{\left | C_a \right |}\sum_{j \in C_a , i \neq j}d\left ( i,j \right )
		\end{equation*}\\
		
		\item
        For each example, the minimum average distance $b(i)$ between the example and the examples contained in each cluster not containing the analyzed example:
		
		\begin{equation*}
			b(i)= \min_{C_b \neq C_a} \frac{1}{\left | C_b \right |}\sum_{j \in C_b}d\left ( i,j \right )
		\end{equation*}\\
		
		\item For each example, the silhouette coefficient is determined by the following expression:
		
		\begin{equation*}
			s(i)=\frac{b(i) - a(i)} {\max\{a(i),b(i)\}}
		\end{equation*}\\
		
	\end{enumerate}

	The silhouette coefficient is defined in the interval $[-1,1]$ for each example in our data set. The global silhouette coefficient is just the average of the particular silhouette coefficients for each example:
	
	\begin{equation*}
		S = \frac{1}{n}\sum_{i=1}^{n}s(i)
	\end{equation*}\\
	
	Unlike other combined measures, the silhouette coefficient provides us a simple framework for qualification. Positive values indicate a high separation between clusters. Negative values are an indication that the clusters are mixed with each other (i.e. an indication of overlapping clusters). When the silhouette coefficient is zero, it is an an indication that the data are uniformly distributed throughout the Euclidean space \cite{aggarwal_data_2015}.\\
	
	Unfortunately, one of the main drawbacks of the silhouette coefficient is its high computational complexity, $O(dn^2)$, which makes it impractical when dealing with huge data sets \cite{celebi_unsupervised_2016}.\\

\end{itemize}


Despite their widespread use, cohesion and separation metrics are not the only validation method available for partitional clustering techniques. In fact, cohesion and separation metrics do not perform well when it comes to analyzing results obtained by algorithms based on density analysis.\\

Given the proximity (or similarity) matrix of a data set and the clustering obtained by a clustering algorithm, we can compare the actual proximity matrix to an ideal version of the proximity matrix based on the provided clustering. Reordering rows and columns so that all examples of the same cluster appear together, the ideal proximity matrix has a block diagonal structure. High correlation between the actual and ideal proximity matrices indicates that examples in the same cluster are close to each other, although it might not be a good measure for density-based clusters \cite{tan_introduction_2005}.\\

Unfortunately, the mere construction of the whole proximity matrix is computationally expensive, $O(n^2)$, and this validation method cannot be used without sampling for large data sets.\\

\subsection{Hierarchical Methods}

The clustering validation methods discussed in the previous section were devised for partitional clustering algorithms. Several internal validation techniques have also been proposed and tested with hierarchical clustering algorithms. As you can expect, these evaluation metrics obtain better results when using hierarchical algorithms such as the single link agglomerative clustering algorithm, SLINK \cite{gan_data_2007}.\\

\subsubsection{Cophenetic Correlation Coefficient (CPCC)}
	
The cophenetic distance between two examples is the proximity at which an agglomerative hierarchical clustering algorithm puts the examples in the same cluster for the first time \cite{tan_introduction_2005}. Looking at the associated dendrogram, it corresponds to the height at which the branches corresponding to the two examples are merged.\\

The cophenetic correlation coefficient (CPCC) is a metric used to evaluate the results of a hierarchical clustering algorithm \cite{gan_data_2007}. This correlation coefficient was proposed by Sokal and Rohlf  in 1962 \cite{sokal_comparison_1962} as the correlation between the entries of the cophenetic matrix $P_c$, containing cophenetic distances, and the proximity matrix $P$, containing similarities.\\

The cophenetic matrix $P_c$ defined for pairs of examples $P_c(i,j)$ as the level of proximity between the examples $(i, j)$ in the dendrogram (i.e. the level of proximity at which both examples are assigned to the same cluster). The cophenetic correlation coefficient is then defined as \cite{gan_data_2007}

	\begin{equation*}
	CPCC=\frac{\sum_{i<j}(d_{ij}-\bar{d})(d_{ij}^{*}-\bar{d^{*}})}{\sqrt{\sum_{i<j}(d_{ij}-\bar{d})^2\sum_{i<j}(d_{ij}^{*}-\bar{d^{*}})^2}}
	\end{equation*}\\
	
where $d_{ij}$ is the distance between the example pair $(i, j)$ and $d_{ij}^{*}$ is their cophenetic distance. The correlation coefficient also includes the average $\bar{d}$ of the distances in the proximity matrix and the average $\bar{d^{*}}$ of the cophenetic distances in the cophenetic matrix, which can be computed as follows:
	
	\begin{equation*}
		\bar{d} = \frac{ \sum_{i<j}d_{ij} }{ 2\left(n^2-n\right) }
	\end{equation*}
	
	\begin{equation*}
		\bar{d^{*}} = \sqrt{\frac{\sum_{i<j}\left(d_{ij}-d_{ij}^{*}\right)^2}{\sum_{i<j}(d_{ij}^{*})^2}}
	\end{equation*}\\

The cophenetic correlation coefficient, as the silhouette coefficient and any other correlation coefficient, is a value in the interval $[-1,1]$. High CPCC values indicate a high level of similarity between the two matrices \cite{gan_data_2007}, an indication that the clustering algorithm has been able to identify the underlying structure of its input data.\\

\subsubsection{Hubert Statistic}
	
The Hubert statistic is similar to the cophenetic correlation coefficient. First, concordance are discordance are defined for pairs of examples.\\

A pair $(i,j)$ is concordant when $\left(\left(v_{p_i}<v_{c_i}\right) \& \left(v_{p_j}<v_{c_j}\right)\right)$ or $\left(\left(v_{p_i}>v_{c_i}\right) \& \left(v_{p_j}>v_{c_j}\right)\right)$. Likewise, a pair $(i,j)$ is said to be discordant when $\left(\left(v_{p_i}<v_{c_i}\right) \& \left(v_{p_j}>v_{c_j}\right)\right)$ or $\left(\left(v_{p_i}>v_{c_i}\right) \& \left(v_{p_j}<v_{c_j}\right)\right)$. Therefore, a pair is neither concordant nor discordant if $v_{p_i}=v_{c_i}$ or $v_{p_j}=v_{c_j}$.\\

 Let $S_{+}$ and $S_{-}$ be the number of concordant and discordant pairs, respectively. Then, the Hubert coefficient is defined as \cite{theodoridis_pattern_2003}:
	
	\begin{equation*}
	\gamma = \frac{S_{+} - S_{-}}{S_{+} + S_{-}}\
	\end{equation*}
	
As the cophenetic coefficient, the Hubert statistic is between -1 and 1. Like CPCC, it has been mainly used to compare the results of two hierarchical clustering algorithms. A higher Hubert $\gamma$ value corresponds to a better clustering of data.\\

\section{External Validation}

External validation methods can be associated to a supervised learning problem. External validation proceeds by incorporating additional information in the clustering validation process, i.e. external class labels for the training examples. Since unsupervised learning techniques are primarily used when such information is not available, external validation methods are not used on most clustering problems. However, they can still be applied when external information is available and also when you generate synthetic data from a real data set.\\

\begin{figure}[htbp]
		\centerline
		{
			\includegraphics[scale=0.4]{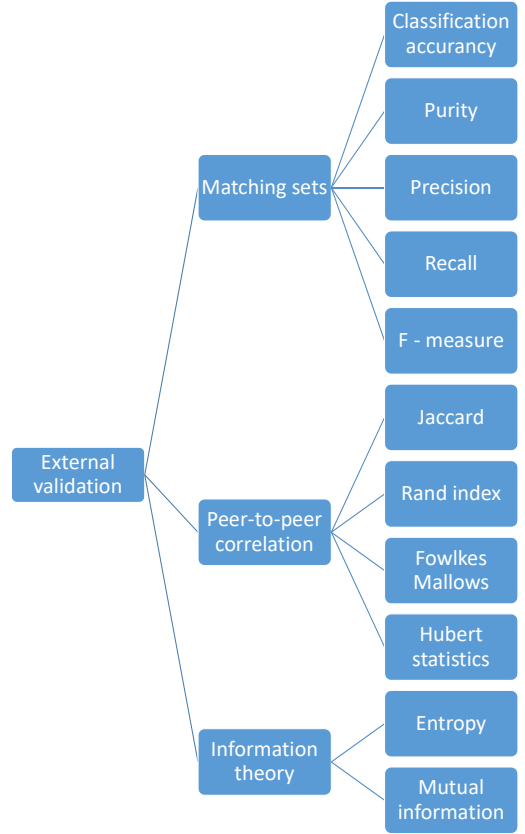}
		}
		\caption{External validation methods (adaptation)\cite{cord_machine_2008}.}
		\label{figura4}
	\end{figure}

Like internal validation methods, it is also possible to classify external metrics depending on the algorithmic approach of the clustering technique used to solve a particular clustering problem. A more rational classification of external validation methods is shown in Figure \ref{figura4} \cite{cord_machine_2008}. According to this taxonomy, different external validation metrics can be used to compare two sets of clusters, the first one obtained by the clustering algorithm under evaluation and the second one provided by an external source.\\
	
We want to compare the result of a clustering algorithm $C=\{C_1,C_2,...,C_m\}$ to a potentially different partition of data $P=\{P_1,P_2,...,P_s\}$, which might represent the expert knowledge of the analyst (his experience or intuition), prior knowledge of the data in the form of class labels, the results obtained by another clustering algorithm, or simply a grouping considered to be ``correct'' \cite{gan_data_2007}.\\
	
In order to carry out this analysis, a contingency matrix must be built to evaluate the clusters detected by the algorithm. This contingency matrix contains four terms:\\
	
	\begin{itemize}
		\item $TP$: The number of data pairs found in the same cluster, both in $C$ and in $P$.\\
		\item $FP$: The number of data pairs found in the same cluster in $C$ but in different clusters in $P$.\\
		\item $FN$: The number of data pairs found in different clusters in $C$ but in the same cluster in $P$.\\
		\item $TN$: The number of data pairs found in different clusters, both in $C$ and in $P$.\\
	\end{itemize}

From these four indicators, we can easily obtain:\\
	
	\begin{itemize}
		\item The number of pairs found in the same cluster in $C$: $m_1=TP+FP$.\\
		\item The number of pairs found in the same cluster in $P$: $m_2=TP+FN$.\\
	\end{itemize}

Obviously, the total number of pairs must be
	
	\begin{equation*}
		M = TP+FP+FN+TN = \frac{n \left( n-1 \right) }{2}
	\end{equation*}\\

\subsection{Matching Sets}
	
The first family of external validation methods that can be used to compare two partitions of data consists of those method that identify the relationship between each cluster detected in $C$ and its natural correspondence to the classes in the reference result defined by $P$.\\
	
Several measures can be defined to measure the similarity between the clusters in $C$, obtained by the clustering algorithm, and the clusters if $P$, corresponding to our prior (external) knowledge \cite{aggarwal_data_2016}:\\

\begin{itemize}

\item
\textbf{Precision} counts the true positives, how many examples are properly classified within the same cluster \cite{perry_machine_1955}:
	\begin{equation*}
		Pr=\frac{TP}{TP+FP}=\frac{TP}{P}=\frac{p_{ij}}{p_i}
	\end{equation*}\\

\item
\textbf{Recall} evaluates the percentage of elements that are properly included in the same cluster:
	\begin{equation*}
		R=\frac{TP}{TP+FN}=\frac{p_{ij}}{p_j}
	\end{equation*}	\\

\item
The \textbf{F-measure} combines precision and recall in a single metric, their weighted harmonic mean:
	\begin{equation*}
		F_{\alpha}=\frac{1+\alpha}{\frac{1}{Pr}+\frac{\alpha}{R}}
	\end{equation*}	\\
Quite often, precision and recall are evenly combined with an unweighted harmonic mean ($\alpha=1$):
	\begin{equation*}
		F=\frac{2*Pr*R}{Pr+R} =\sum_{j}p_j \max_i \left[\frac{2*Pr*R}{Pr+R}\right]
	\end{equation*}\\

\item
\textbf{Purity} evaluates whether each cluster contains only examples from the same class:
	\begin{equation*}
		U = \sum_{i} p_i \left( \max_j \frac{p_{ij}}{p_i} \right)
	\end{equation*}	\\

\end{itemize}

In the expressions above, $p_i=n_i/n$, $p_j=n_j/n$, and $p_{ij}=n_{ij}/n$, where $n_{ij}$ is the number of examples belonging to the class $i$ found in the cluster $j$ and $n_i$ ($n_j$) is the number of examples in the cluster $i$ ($j$).\\

\subsection{Peer-to-peer Correlation}
	
A second family of measures for external validation are based on the correlation between pairs, i.e. they seek to measure the similarity between two partitions under equal conditions, such as the result of a grouping process for the same set, but by means of two different methods $C$ and $P$. It is assumed that the examples that are in the same cluster in $C$ should be in the same class in $P$, and vice versa \cite{tan_introduction_2005}.\\

Some metrics based on measuring the correlation between pairs are the following:\\

\begin{itemize}

\item
The \textbf{Jaccard coefficient} assesses the similarity of the detected clusters $C$ to the provided partition $P$:
	\begin{align*}
		J & = \frac{TP}{TP+FP+FN} \\
        J & =\frac{\sum_{ij}\binom{n_{ij}}{2}}{ \sum_i\binom{n_i}{2}+\sum_{j}\binom{n_j}{2}-\sum_{ij}\binom{n_{ij}}{2}}
	\end{align*}\\

\item
The \textbf{Rand coefficient} is similar to the Jaccard coefficient, yet it is measured against the total data set (equivalent to accuracy in a supervised machine learning setting):
	\begin{align*}
		Rand & = \frac{TP+TN}{M} \\
        Rand & = \frac{ \binom{n}{2}-\sum_i\binom{n_i}{2}+\sum_{j}\binom{n_j}{2}-\sum_{ij}\binom{n_{ij}}{2} }{\binom{n}{2}}
	\end{align*}\\
	
\item
The \textbf{Folkes and Mallows coefficient} computes the similarity between the clusters found by the algorithm with respect to the independent markers:
	
	\begin{align*}
		FM & = \sqrt{\frac{TP}{TP+FP}*\frac{TP}{TP+FN}} \\
        FM & = =\frac{\sum_{ij}\binom{n_{ij}}{2}}{\sqrt{\sum_i\binom{n_i}{2}*\sum_{j}\binom{n_j}{2}}}
	\end{align*}\\

\item
We can also resort to the \textbf{Hubert statistical coefficient} in this context:
	
	\begin{align*}
	\Gamma & = \frac{1}{M}\sum_{i=1}^{n-1}\sum_{j=i+1}^{n}X_{ij}Y_{ij} \\
    \Gamma & = \frac{ \binom{n}{2}\sum_{ij}\binom{n_{ij}}{2}-\sum_{i}\binom{n_i}{2}\sum_{j}\binom{n_j}{2} } { \sqrt{\sum_{i}\binom{n_i}{2}\sum_{j} \binom{n_j}{2} \left[\binom{n}{2}-\sum_{i}\binom{n_i}{2}\right] \left[\binom{n}{2}-\sum_{j}\binom{n_j}{2}\right]} }
	\end{align*}\\

	
	

\end{itemize}

As before, $n_{ij}$ is the number of examples belonging to the class $i$ found in the cluster $j$, whereas $n_j$ ($n_j$) is the number of examples in the cluster $i$ ($j$).\\

\subsection{Measures Based on Information Theory}

A third family of external cluster validation metrics is based on Information Theory concepts, such as the existing uncertainty in the prediction of the natural classes provided by the partition $P$. This family includes basic measures such as entropy and mutual information, as well as their respective normalized variants.\\

\begin{itemize}

\item
\textbf{Entropy} is a reciprocal measure of purity that allows us to measure the degree of disorder in the clustering results:
	\begin{equation*}
		H = -\sum_{i}p_i\left(\sum_{j}\frac{p_{ij}}{p_i}\log{\frac{p_{ij}}{p_i}}\right)
	\end{equation*}\\

\item	
\textbf{Mutual information} allows us to measure the the reduction in uncertainty about the clustering results given knowledge of the prior partition:
	\begin{equation*}
		MI = \sum_{i}\sum_{j} p_{ij} \log\frac{p_{ij}}{p_ip_j}
	\end{equation*}

\end{itemize}

As always, $p_{ij}=n_{ij}/n$, $p_i=n_i/n$, and $p_j=n_j/n$.\\



\section{Hyperparameter Tuning}
	
Internal and external validation metrics are used once the clustering algorithm has been applied to the available data set. However, the clustering algorithm itself has its own parameters. Adjusting those parameters, also known as hyperparameters in the machine learning literature, can help us obtain very different clustering results.\\

When using unsupervised machine learning techniques, several issues affect their effectiveness. Even though external validation metrics can help us evaluate whether the obtained clusters closely match the underlying categories in the training data, which the clustering algorithm tries to identify without externally-provided class labels, those metrics cannot address other issues such as the right number of clusters for our current data set. For instance, in the case of hierarchical clustering techniques, we are certainly interested in determining the best level at which we can cut our dendrogram.\\

Any clustering algorithms has a set of parameters $P_{alg}$, which might include the number of clusters $n_c$ or not. Hyperparameter tuning tries to determine, for the different possible values of the parameters in $P_{alg}$, which set of parameter values is the most suitable for our particular clustering problem.\\
 
We could proceed in the following way \cite{halkidi_clustering_2001}:\\

\begin{itemize}

\item 
When the algorithm does not include the number of clusters $n_c$ among its parameters ($n_c \notin P_{alg}$), we run the algorithm with different values for its parameters so that we can determine their largest range for which $n_c$ remains constant. Later, we choose as parameter values the values in the middle of this range.\\

\item
When the algorithm parameters $P_{alg}$ include the desired number of clusters $n_c$ ($n_c \in P_{alg}$), we run the algorithm for a range of values for $n_c$ between $n_{cmin}$ and $n_{cmax}$. For each value of $n_c$, we run the algorithm multiple times using different sets of values (i.e. starting from different initial conditions) and choose the value that optimizes our desired validation metric, which might be internal or external depending on our particular clustering problem.\\

\end{itemize}

When we just want to determine the ``right'' number of clusters, $n_c$, plotting the validation results for different values of $n_c$ can sometimes show a relevant change in the validation metric, commonly referred to as a ``knee'' or ``elbow'' \cite{theodoridis_pattern_2003}.\\

However, in practice, the number of parameters in $P_{alg}$ might be large, so we cannot test every possible value combination of parameter values in a systematic way. Hyperparameter tuning can then be seen as a combinatorial optimization problem. Fortunately, we can resort to automated tuning strategies that facilitate our search \cite{Bergstra:2012:RSH:2188385.2188395}. Among the tuning strategies at our disposal, we could include the following ones \cite{berzal_redes_2019}:\\
	
\begin{itemize}
	
\item 
\textbf{Grid search} is based on a systematic exploration of the hyperparameter space, on which we define a grid and test for a parameter combination in each cell of such grid. If we have $p$ parameters, the systematic exploration of a $p$-dimensional space might require an exponential number of parameter configurations, often unfeasible in practice.\\

\item
\textbf{Random search} chooses parameter configurations at random. Even though our search is not exhaustive, when using random search, we hope that some parameter combinations will lead us to promising regions in our search space. The rationale behind random search is that, quite often, local changes in the parameter values do not produce significant changes in the algorithm output, so that a systematic exploration might not be really necessary, even when feasible (usually, it is not feasible anyway).\\
		
\item 
\textbf{Smart search} techniques try to optimize the problem of searching for hyperparameter values. Different strategies can be implemented, such as Bayesian optimization using Gaussian processes and evolutionary optimization using genetic algorithms or evolution strategies. \\

\end{itemize}

\section{Conclusion}
	
Determining the quality of the results provided by a clustering algorithm is not an easy problem. Kleinberg defined three properties any clustering algorithm should try to satisfy (the axioms of scale invariance, richness, and consistency) and proved an impossibility theorem that shows that no clustering algorithm can simultaneously satisfy all of them.\\

A wide range of metrics have been proposed in the literature to quantify the quality of clustering results:\\
 
\begin{itemize}

\item
Internal validation metrics do not require external information. These metrics focus on measuring cluster cohesion and separation, on the statistical analysis of the proximity matrix, or on the study of the dendrogram generated by hierarchical clustering algorithms.\\

\item
External validation metrics	resort to externally-provided information to evaluate the quality of the clustering results. A large number of external validation metrics are at our disposal, ranging from matching sets to peer-to-peer correlation and information theoretical indices. \\

\end{itemize}

External validation metrics are also useful when comparing the results provided by different clustering algorithms (or the same algorithm with different sets of parameter values). \\

Finding the best configuration for the parameters of an algorithm is known as hyperparameter tuning. This process is often necessary, for instance, to determine the optimal number of clusters for a particular clustering problem.\\

	\bibliographystyle{ieeetr}
	\bibliography{biblio}
	
\end{document}